\pdfoutput=1

\documentclass[11pt]{article}

\usepackage{ACL2023}

\usepackage{times}
\usepackage{latexsym}

\usepackage[T1]{fontenc}

\usepackage[utf8]{inputenc}

\usepackage{microtype}

\usepackage{inconsolata}

\usepackage{amsthm}
\usepackage{amsmath}
\usepackage{amsfonts}
\usepackage{amssymb}
\usepackage{caption}
\usepackage{subcaption}
\usepackage{dsfont}
\usepackage{mathtools}
\usepackage{bbm}
\usepackage{bm}
\usepackage{multirow}
\usepackage{multicol}
\title{Are fairness metric scores enough to assess discrimination biases in machine learning?}

\author{Fanny Jourdan \\
  IRIT, Université Paul-Sabatier  \\
  Toulouse, France \\
  \texttt{fanny.jourdan@irit.fr }
  \And
  Laurent Risser \\
  IMT, Université Paul-Sabatier  \\
  Toulouse, France \\ \AND
  Jean-Michel Loubes \\
  IMT, Université Paul-Sabatier  \\
  Toulouse, France \\ \And
  Nicholas Asher \\
  IRIT, Université Paul-Sabatier  \\
  Toulouse, France \\
  }

\begin{document}
\maketitle
\begin{abstract}
This paper presents novel experiments shedding light on the shortcomings of current metrics for assessing biases of gender discrimination made by machine learning algorithms on textual data. We focus on the \emph{Bios} dataset, and our learning task is to predict the occupation of individuals, based on their biography. Such prediction tasks are common in commercial Natural Language Processing (NLP) applications such as automatic job recommendations. 
We address an important limitation of theoretical discussions dealing with group-wise fairness metrics: they focus on large datasets, although the norm in many industrial NLP applications is to use small to reasonably large linguistic datasets for which the main practical constraint is to get a good prediction accuracy. We then question how reliable are different popular measures of bias when the size of the training  set is simply sufficient to learn reasonably accurate predictions.
Our experiments sample the \emph{Bios} dataset and learn more than 200 models on different sample sizes. This allows us to statistically study our results and to confirm that common gender bias indices provide diverging and sometimes unreliable results when  applied to relatively small training and test samples.  This highlights the crucial importance of variance calculations for providing sound results in this field.
\end{abstract}

\section{Introduction}

Potential biases introduced by  Artificial Intelligence (AI) systems are now both an academic concern, but also a critical problem for industry, as countries plan to regulate AI systems that could adversely affect individual users.  The so-called \emph{AI act}\footnote{\url{https://eur-lex.europa.eu/legal-content/EN/TXT/HTML/?uri=CELEX:52021PC0206&from=EN}} will require AI systems sold in the European Union to have good statistical properties with regard to any potential discrimination they could engender.  In particular, under the AI Act, AI systems that exploit linguistic data like those for reviewing job candidates from text-based candidacies fall into the category of tightly regulated AI systems, as they are intended to be used for the recruitment or selection of natural persons (see Annex III of the AI act).  Such AI systems will require frequent and rigorous statistical testing for unwanted biases.\footnote{Such AI systems are considered high-risk. The {\em AI act} (Article 9.7) states: "\emph{The testing of the high-risk AI systems shall be performed, as appropriate, at any point in time throughout the development process, and, in any event, prior to the placing on the market or the putting into service. Testing shall be made against preliminarily defined metrics and probabilistic thresholds that are appropriate to the intended purpose of the high-risk AI system}". Article 10.2 specifies that "\emph{Training, validation, and testing data sets shall be subject to appropriate data governance and management practices. Those practices shall concern in particular, examination in view of possible biases}" (among others). Article 71 states that "\emph{non-compliance of the AI system with the requirements laid down in Article 10 ... shall be subject to administrative fines of up to 30 000 000 EUR or, if the offender is a company, up to 6 \% of its total worldwide annual turnover for the preceding financial year}".}

These regulatory advances have made it a pressing issue to define which  metrics are appropriate for evaluating whether machine learning models can be considered fair algorithms in various industrial settings. In this context, we believe that these articles open at least two issues:
(1) Each fairness metric quantifies the fairness of a model in a different way and not all metrics are compatible with each other, as already discussed in \cite{kleinberg2016inherent, chouldechova2017fair, pleiss2017fairness}. It is therefore easy to optimize its algorithm according to a single metric to claim fairness while overlooking all the other aspects of fairness measured by other metrics.
(2) Given that contemporary, theoretical discussions of fairness focus on large datasets but that the norm in many industrial NLP applications is to use small linguistic datasets \cite{ezen2020comparison}, one can wonder how reliable different popular measures of bias when the size of the training and validation sets is simply sufficient to learn reasonably accurate predictions. 
In general, this leads us to pose two questions, which are central to this paper: 
Are fairness metrics always reliable on small samples, which are common in industrial contexts? How do they behave when applying standard debiasing techniques?


To answer these questions, we propose a new experimental protocol to expose gender biases in NLP strategies, using variously sized subsamples of the \textit{Bios} dataset \cite{de2019bias}.  We create 50 samples for each sample size (10k, 20k, 50k, and 120k) and train a model on each of the 200 samples. This gives us a mean and a variance on our results for all sample sizes to be able to compare them from a statistical point of view.
We study the biases in these samples  using three metrics; each sheds light on specific properties of gender bias. 

Our study shows how bias is related to the training set size on a standard NLP dataset by revealing three points:  
First, commonly accepted bias indices appear unreliable when computed on ML models trained on relatively small training sets. Moreover, our experiments reveal that the group parity gender gap metric (\ref{metric}) 
appears to be more reliable than other metrics on small samples.  
Second, in the tested standard and large training sets, results are not homogeneous across professions and across the measures: sometimes there is gender bias against males, and sometimes against females in professions where one would expect something different.  Finally, the most traditional de-biasing methods, which consist in removing gender-sensitive words or replacing them with neutral variants, makes different metrics yield surprising and sometimes seemingly incompatible bias effects. We explain this phenomenon by the definitions of the metrics. In light of these findings, we think that one should use the main fairness metrics jointly to look for biases in smaller datasets and run enough models to have a variance. Such bootstrapping procedures appear essential to robustly analyze how fair a prediction model is.

Our paper is structured as follows.  Section 2 surveys related work. Section 3 introduces our experimental setup.  Section 4 discusses our results, with conclusions coming in Section 5. Section 6 discusses some of the limitations of our work.

\section{Related Work}

Gender bias is pervasive in NLP applications: in machine translation \citep{vanmassenhove2019getting, stanovsky2019evaluating, savoldi2021gender, wisniewski2021biais}, in hate speech detection \citep{park2018reducing, dixon2018measuring}, sentiment analysis \citep{kiritchenko2018examining, zellers2019hellaswag}, and in coreference resolution \citep{rudinger2018gender, zhao2018gender}. Gender bias with respect to classification has already been examined in \cite{de2019bias, gonen2019lipstick, bolukbasi2016quantifying, lu2020gender, bordia2019identifying}, and reduced in \cite{pruksachatkun2021robustness, zhao2019gender, zhao2017men}. In particular, for the BERT model,  \citet{bhardwaj2021investigating} investigated gender bias. 
More generally, \citet{bender2021dangers} has studied the impact of increasingly large language models and has highlighted the sexist or racist biases and prejudices that result from them.

However, the above-mentioned works only focused on single, large datasets. 
Recently, a growing literature has started to propose to leverage statistical properties of fairness metrics, thus providing both sophisticated analysis and practically useful algorithms~\cite{lum2022biasing, diciccio2020evaluating}. In particular, a more rigorous statistical approach for BERT models was introduced in~\cite{sellam2021multiberts}.

In this paper, we investigate the pertinence of different fairness metrics on samplings of different sizes out of a large dataset. We apply our principled statistical procedure and we present the results of these measures, along with their standard deviation and properties coming from Student's t-tests. 
In addition to our scientific contribution, we have paid particular attention to the clarity of our explanations and the simplicity of our proposed protocol to allow small players to easily employ them for their real-world use cases.
Finally, our results attest to the importance of applying techniques of statistical analysis to Fairness problems, and we hope that the guarantees gained through them provide a convincing argument for its more generalized application in the field.

\section{Experimental protocol}

In this section, we detail the various components of our experimental setup.  Section 3.1 describes the dataset and  Section 3.2 the general type of model used to train the 200 models.  Section 3.3 introduces our debiasing technique used to illustrate our protocol. Section 3.4 explains the sampling procedure and gives guarantees on the representativeness of the samples. Finally, Section 3.5 describes the different fairness metrics that we will compare and we justify these choices.


\subsection{The {\em Bios} data set}
The \textit{Bios} dataset \cite{de2019bias} contains about 400K biographies (textual data). 
For each biography, \textit{Bios} specifies the gender (M or F) and the occupation (among 28 occupations, categorical data) of its author. Figure \ref{fig:distribution} (Appendix) shows the distribution of each occupation by gender.

\subsection{DistilBERT model}
Our task is to predict the occupation using only the textual data of the biography.  This task is relevant in the case of our study because job prediction from LinkedIn biographies is used for job recommendation. It is therefore easy to imagine the consequences of gender discrimination in this context.


For this task, we will use the DistilBERT architecture.  DistilBERT \citep{sanh2019distilbert} is a transformer architecture derivative from but smaller and faster than the original BERT \citep{devlin2018bert}. 
This model is commonly used to do text classification. DistilBERT is trained on BookCorpus \cite{moviebook} (like BERT), a dataset consisting of 11,038 unpublished books and English Wikipedia (excluding lists, tables and headers), using the BERT base model as a teacher. 

We have fine-tuned  DistilBERT  to adapt it to our text classification task. 
In our protocol, only the datasets were intervened on while keeping other factors the same in each model.
We used 5 epochs,  a batch size of 16 observations, an AdamW optimizer with a learning rate of 2e-5, and a cross-entropy loss when training the  model. 

\subsection{De-biasing methodology}

In this part, we state the debiasing technique used for the illustration of our protocol. Note that this technique is very basic and is only used to explain our experimental protocol. This protocol could be applied with any more elaborate debiasing technique.

A classic method for debiasing consists of removing explicit gender indicators ({\em i.e.}{\em 'he', 'she', 'her', 'his', 'him', 'hers', 'himself', 'herself', 'mr', 'mrs', 'ms', 'miss'} and first names).
For a model like DistilBERT, however, we could not just remove words because the model is sensitive to sentence structure, not just lexical information. 
We, therefore, adjusted the method by replacing all the first names with a neutral first name\footnote{We can take any first name because, since we change all the first names of the dataset by this one, it will necessarily be neutral.} (\textit{Camille}) and by choosing only one gender for all datasets (e.g., for all individuals of gender g, we did nothing; for the others, we replaced explicit gender indicators with those of g).
We then created two datasets with only female or male gender indicators, and the only first name \textit{Camille}.

\subsection{Sampled training and test sets}
We tested the robustness of our model with respect to the various bias measures on training sets of different sizes. We randomly sampled 50 different training sets containing 10K, 20K, 50K, and 120K biographies out of the 400K of \cite{de2019bias}.
We trained a model on each of these 200 samples. Each of these models has the same architectures and the same hyper-parameters stated previously. To guarantee the representativeness of the sample, we ensured that each sample had the same percentage of each gender for each occupation as in the initial data set.  For example, given 2002 female surgeons out of 388862 persons in the initial dataset (0.51 \%), we  randomly picked 51 women surgeons for a sample with 10000 individuals (0.51 \%). For the split between the train and test sets, we  respectively used 70\% and 30\% of the dataset. 

Creating these 200 different models and observations makes it possible to quantify the variability of the results obtained using each size of subsampled training sets. This will additionally allow us to ensure that all differences discussed in our results are statistically significant using Student's t-tests.
Our experimental protocol, therefore, gives us more guarantees than traditional protocols based on a single model.

\subsection{Gender bias metrics}\label{metric}

Let $\hat{Y}$ and $Y$ be the predicted and the true target labels (\textit{i.e.}, the occupations), respectively. Let $G$ be a random variable representing the binary gender of the biography’s subject. For each model, we quantified the gender bias by using the following metrics: 
Group Parity (GP), True Positive Rate (TPR), and Predictive parity (PP). They are  defined as:

\begin{eqnarray}
  GP_{g,y} = P(\hat{Y} = y | G = g) \,,\\
  TPR_{g,y} = P(\hat{Y} = y | G = g, Y=y) \,,\\
  PP_{g,y} = P(Y = y | \hat{Y} = y, G = g) \,.
\end{eqnarray}

To measure the gender gap with these metrics, we computed the difference between binary genders $g$ and $\tilde g$ — for each occupation $y$:

\begin{equation}\nonumber
    M\_Gap_{g,y} = M_{g,y} - M_{\tilde g,y} \,,
\end{equation} 
where $M$ is $GP$, $TPR$ or $PP$.
We now discuss each measure in turn.

\paragraph{Statistical/Group Parity (GP)}
The condition GP verifies gender balance (ie. $GP\_Gap_{g,y} = 0$) if the males and females groups have equal probability of being assigned to the predicted class $\hat{y} \in \{noty,y\}$. It is the most famous and intuitive metric of fairness.

\paragraph{True Positive Rate (TPR)}
$TPR\_Gap_{g,y} = 0$ if among all individuals in the $y$ class, the probability of being predicted $\hat{y}$ is the same for males and females. This metric is widely used in the field of NLP in particular.

\paragraph{Predictive Parity (PP)}
$PP\_Gap_{g,y} = 0$ if among all individuals predicted to belong to the $y$ class, the probability of $Y=y$ is the same for males and females. PP is similar to Calibration (within groups), and widely used in fairness to compare with other metrics. We use PP here because it uses the same parameters $Y$ and $\hat{Y}$ as our other metrics.

\section{Results and discussion}

\begin{figure*}[!h] 
    \centering
    \includegraphics[width=0.9\linewidth]{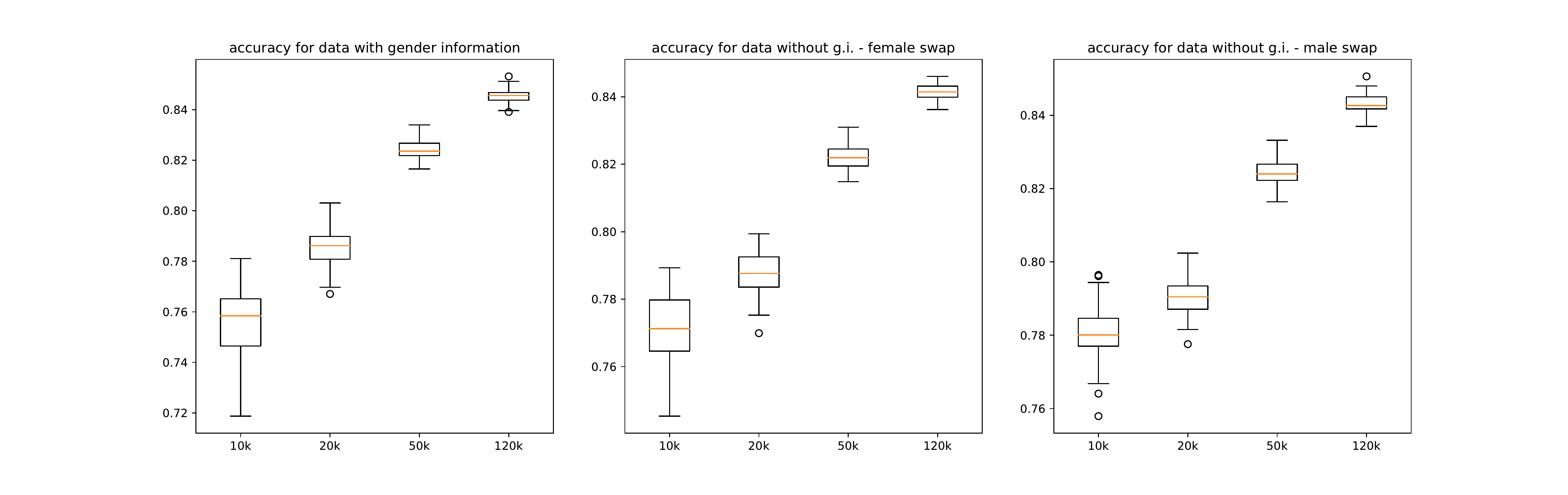}
    \caption{Boxplots representing the variations of prediction accuracy for all sampling sizes}
    \label{fig:Results_accuracies}
\end{figure*}

As shown in Figure~\ref{fig:Results_accuracies}, all the models we trained reached a prediction accuracy ranging from $0.72$ to $0.86$, as in \cite{de2019bias}, which we consider as good since the classification problem involved 28 different occupations.  

All comparisons in this part were considered as significant by using Student's t-tests (p-value of $0.05$).

We created two datasets without gender information: one version with all female gender indicators and the other with all male gender indicators. Gender, therefore, has no impact on the finetuning part of our model. However, since we are starting from a pre-trained DistilBERT model (without a gender-neutral dataset), we had to check that the pre-training had no impact on the prediction. We therefore also made a Student's test between the predictions of one model trained on the dataset with all the female gender indicators, and of another trained on the dataset with all the male gender indicators. The difference was not statistically significant; using one model or the other makes no difference.

The analysis of the results of our protocol is made in two steps: a specific part and a general part.
Below in Section \ref{sec:specific}, we analyze biases on two specific occupations, {\em Surgeon} and {\em Physician}. These two occupations are socially very interesting and their male/female distribution is very different, which is something we wanted to study.  In Section \ref{sec:general}, we also observe the biases across the gamut of occupations in {\em bios}.  All the results found in this preliminary study remain valid in a generalized case where we look at all the classes of the model. \\
Dividing our study like this allows us to discuss various details which support our key message without weighing down the article in the specific part while guaranteeing that our analysis is global and applies to the other classes of the model in the general part.

\subsection{Results and discussion for the classes Surgeon and Physician} \label{sec:specific}

\begin{figure*}[!h] 
    \centering
    \includegraphics[width=0.9\linewidth]{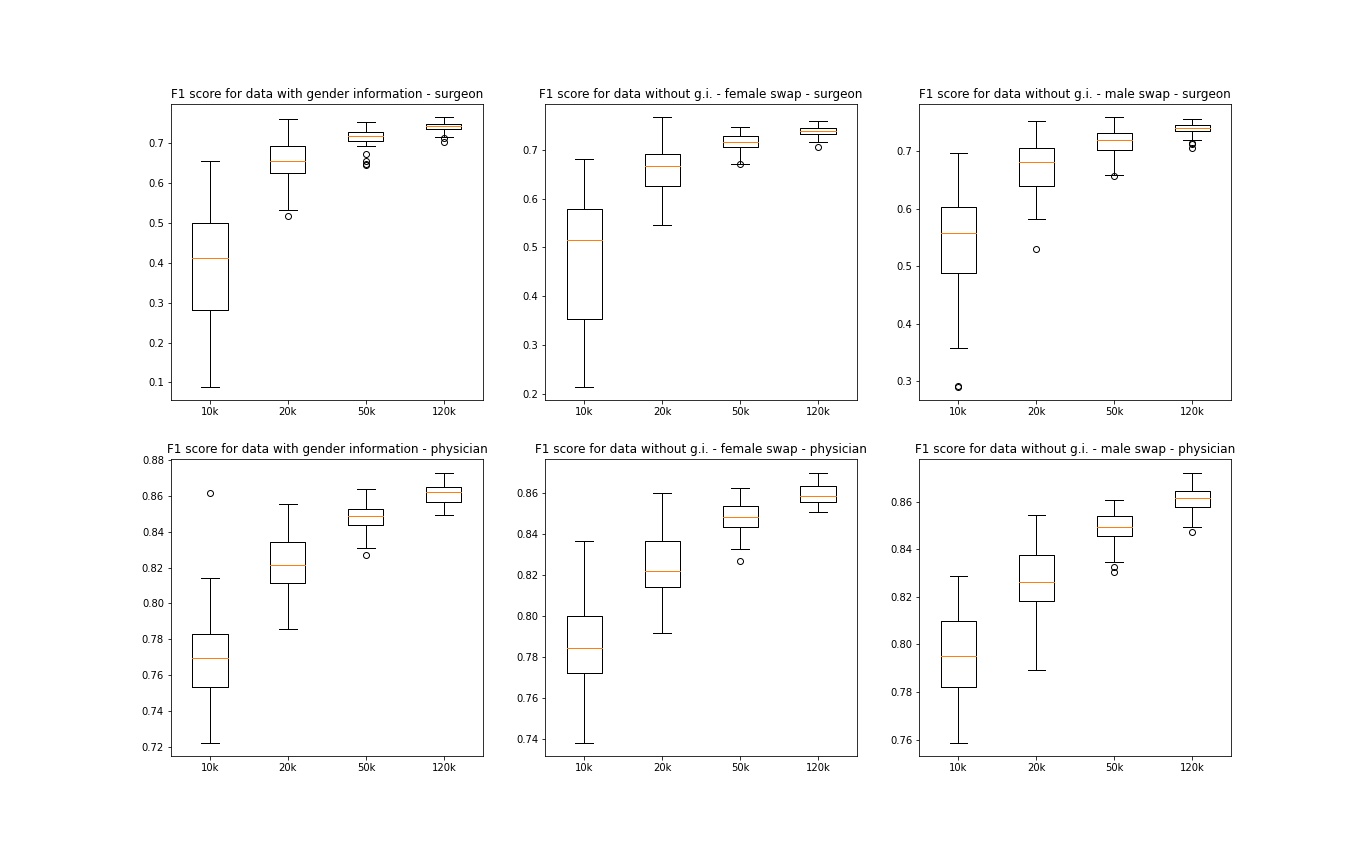}
\caption{Boxplots representing the variations of prediction F1-scores for all sampling sizes for surgeon (top) and physician (bottom)}
    \label{fig:Results_F1score}
\end{figure*}

\begin{figure*}[!h] 
    \centering
        \includegraphics[width=0.9\linewidth]{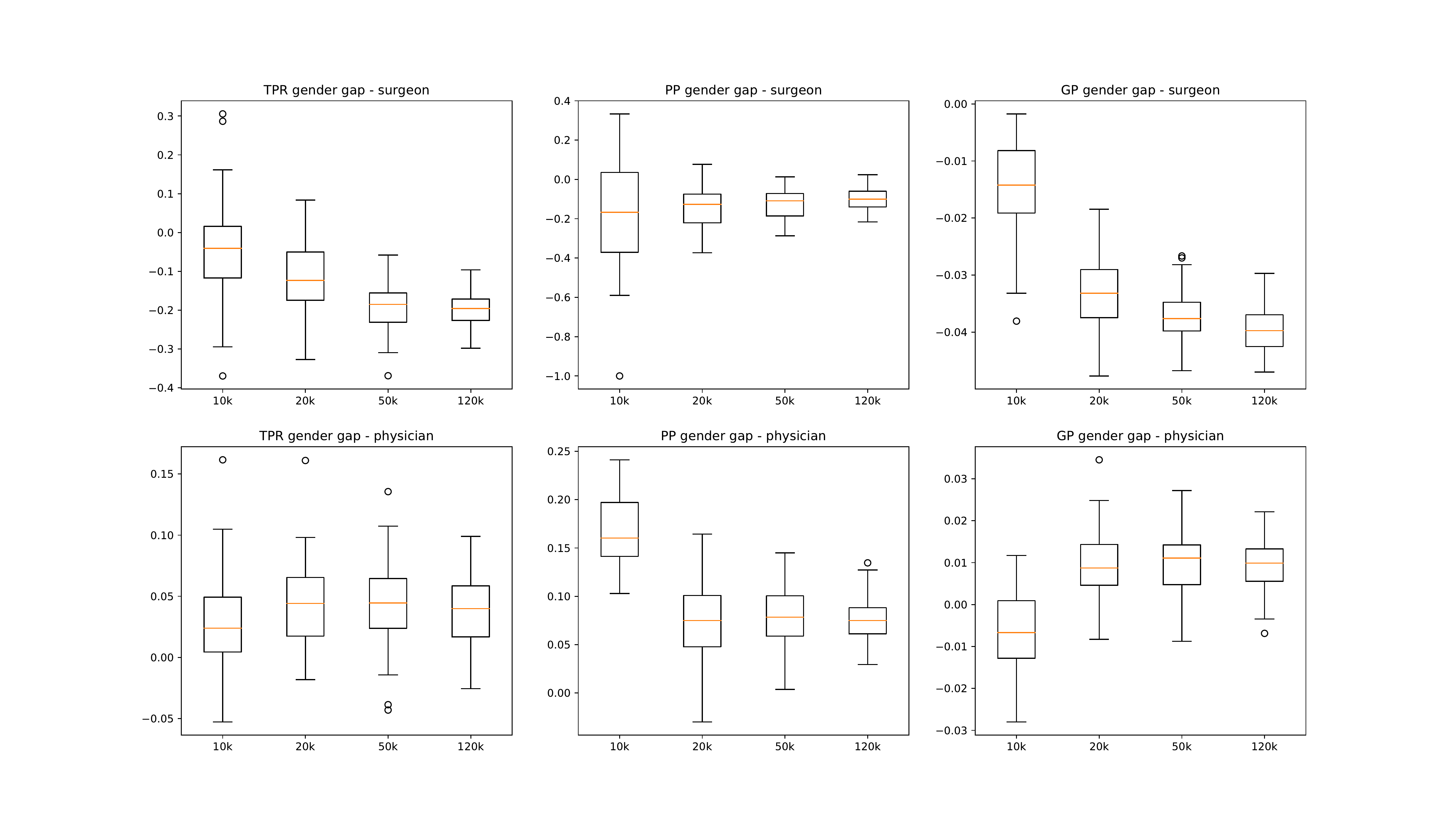}
\caption{Boxplots of the gender gaps obtained using  10K, 20K, 50K, and 120K randomly sampled observations (50). \textbf{(Left)} True Positive Rate (TPR) gender gaps for surgeons and physicians; \textbf{(Middle)} Predictive Parity (PP) gender gaps for surgeons and physicians; \textbf{(Right)} Group Parity (GP) gender gaps for surgeons and physicians.}
    \label{fig:ResultsBERT}
\end{figure*}

Although the model is trained to predict  the occupations of bio authors from the 28 possible choices, we focus, in our study, on the analysis of the biases on two specific occupations:  \emph{Surgeon} versus the 27 remaining occupations, and  \emph{Physician} versus the other occupations. We chose these professions so that we could compare an occupation with an imbalanced gender distribution and one with balanced a gender distribution. The occupation \emph{Physician} is well balanced in the training set between males and females (49,5\% female), while the training set for \emph{Surgeon} contains many more males than females (15\% female).

\begin{figure*}[!h] 
    \centering
    \includegraphics[width=0.9\linewidth]{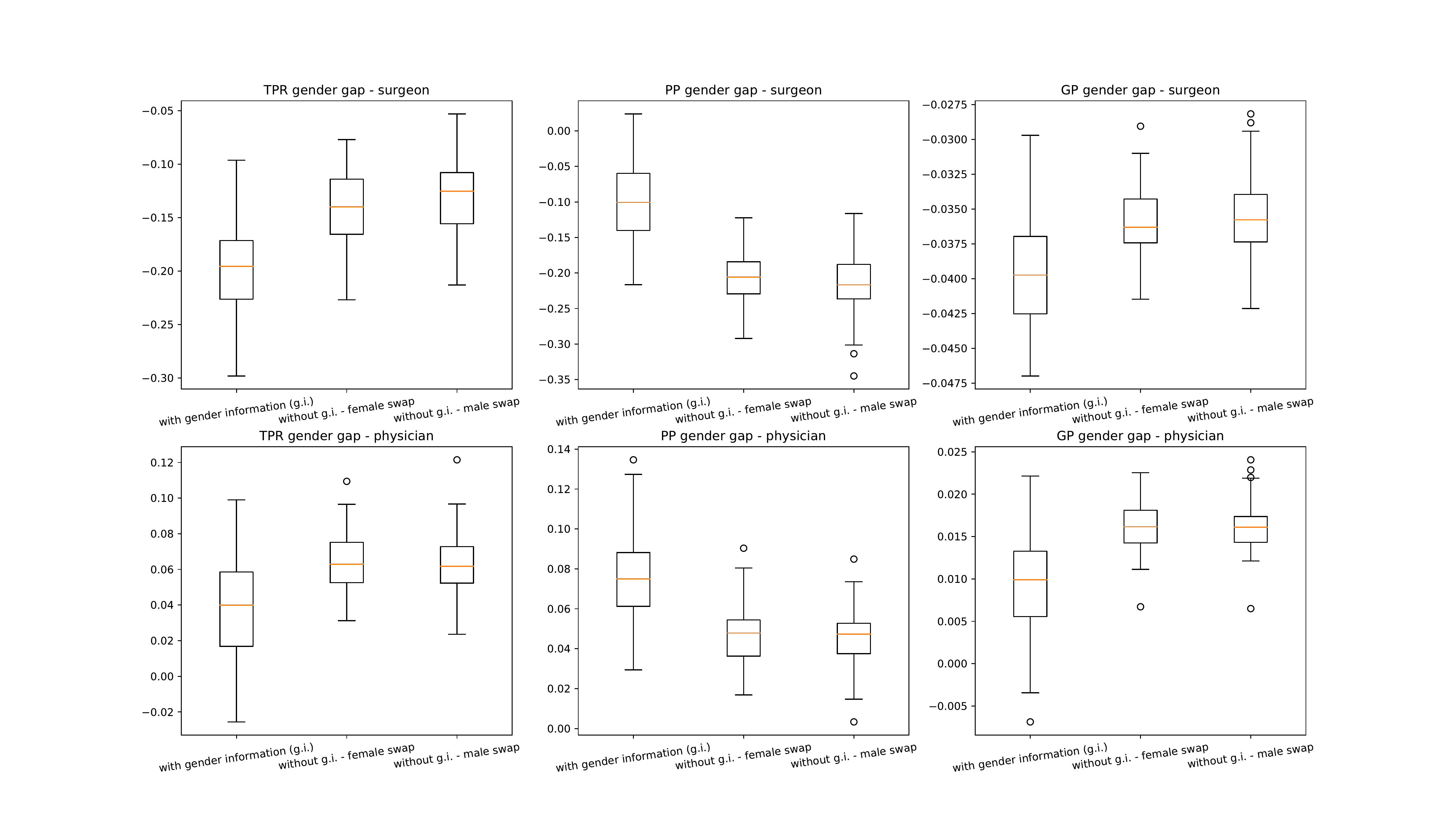}
    \caption{Boxplots representing the fairness metrics for surgeon (top) and physician (bottom) for 120k samplings for the base model, model with only female indicators, and model with only male indicators}
    \label{fig:Results_models}
\end{figure*}

We computed F1-scores in Figure~\ref{fig:Results_F1score}, which are good to reasonable, except for the 10K samples for surgeons, which appear as too small for our predictive task. Quantitative results related to the fairness metrics are shown in Figures \ref{fig:ResultsBERT} and \ref{fig:Results_models}. Each box-plot contains the TPR, GP, PP Gender Gaps obtained on the test set for \emph{surgeons} and \emph{physician}. Negative (positive) gender gaps mean that there is discrimination against females (males). 

\subsubsection{Results on small data samples} 

Our experiments clearly show that the lower the amount of observations in the training set, the more the fairness metrics vary in the test set. The samples with 10K and 20K observations present particularly unstable biases.
%
For example, most TPR (resp. GP) Gender Gaps are negative (resp. positive) for \emph{surgeon} (resp. \emph{physician}) but some samples yield positive TPR (resp. negative GP) Gender Gaps. This is problematic since we cannot deduce a priory that a particular sample should produce discrimination one way or the other.

In addition, the average biases also depend on the sample size. Again, we obtained unstable average biases for small samples (10K, 20K). 
The bias indicators are estimated on the minority class: an amount of 41, 115, 334, and 903 predicted surgeons were obtained in the test set for the 10K, 20K, 50K, and 120K sampling sizes. Hence, their estimation is unstable for small samples. 
However, GP appears more stable than the other metrics in our experiments, in particular when there were few observations. Its variance was indeed close to $0.01$, which is much lower than the variances of $0.1$ and $0.2$ for GP and PP, respectively. We explain this because on our dataset, for TPR and PP, they do not use all predicted surgeons (unlike GP), but only the predicted surgeons who are also real surgeons (in 10k sampling, there are 41 predicted surgeons vs. 30 real surgeons and predicted surgeons, which is an information loss of 26,8\%). 
We explain this intuition mathematically in detail in appendix \ref{apdx:mathintuition}. 

Considering this result, we recommend using a simpler indicator like GP gender gap for small-size sets.

\subsubsection{Bias analysis with different metrics}
\paragraph{General results}
Even for large samples with 120K observations, biases sometimes differed from what we expected.  For the occupation {\em surgeon} (15\% of females) the gender gap was negative for all metrics, which was expected. 
For {\em physician} (49,5\% of females), we also expected to have a negative or zero gender gap (see  \cite{bolukbasi2016man}). However, the gender gaps were  positive for all metrics, which means that the models discriminated against males. 
This example shows that intuitions of model-builders about biases are not always correct and this awareness should influence model construction and testing.

\paragraph{Results with debiasing}

Intuitively, removing explicit gender indicators should reduce the bias \cite{de2019bias}. As shown Figure~\ref{fig:Results_models}, however, our experiments show that this is not necessarily the case. Using  TPR and GP Gender Gaps, we see a bias initially in favor of women (resp. men) and increases (resp. decreases) for the \textit{physician} (resp. \textit{surgeon}) class after debiasing. Removing gender indicators thus favored women in these two occupations. 

PP Gender Gap shows different effects for debiasing: For \textit{physician} (resp. \textit{surgeon}), the initial bias  in favor of women (resp. men) decreases (resp. increases) after debiasing. Removing gender indicators thus favored men in these two occupations.

To explain this phenomenon, we  can remark that removing gender indicators allowed us to predict more women than before in the two professions. The metrics interpret this differently. By definition, $PP_{f,y}= P\left(Y = y | \hat{Y} = y, G = f \right)$  decreases when the number of $\hat{Y}$ increases. In addition,  $TPR_{f,y}= P\left(\hat{Y} = y | Y = y, G = f\right)$  and $GP_{f,y} = P\left(\hat{Y} = y | G = f \right)$  increases when the number of $\hat{Y}$ increases. 

Using either GP/TPR gender gap or PP gender gap amounts to choosing between focusing on the number of people predicted in the discriminated group (parity) or  focusing on the people in the discriminated group who are well predicted (truth). This explains the different interpretations of these indicators.

\subsection{Results  and discussion for all classes}\label{sec:general}

In this section, we confirm our analysis of the specific occupations of {\em Surgeon} and {\em Physician} from a global point of view on all the classes of the model.

\begin{figure*}[!h] 
    \centering
    \includegraphics[width=0.7\linewidth]{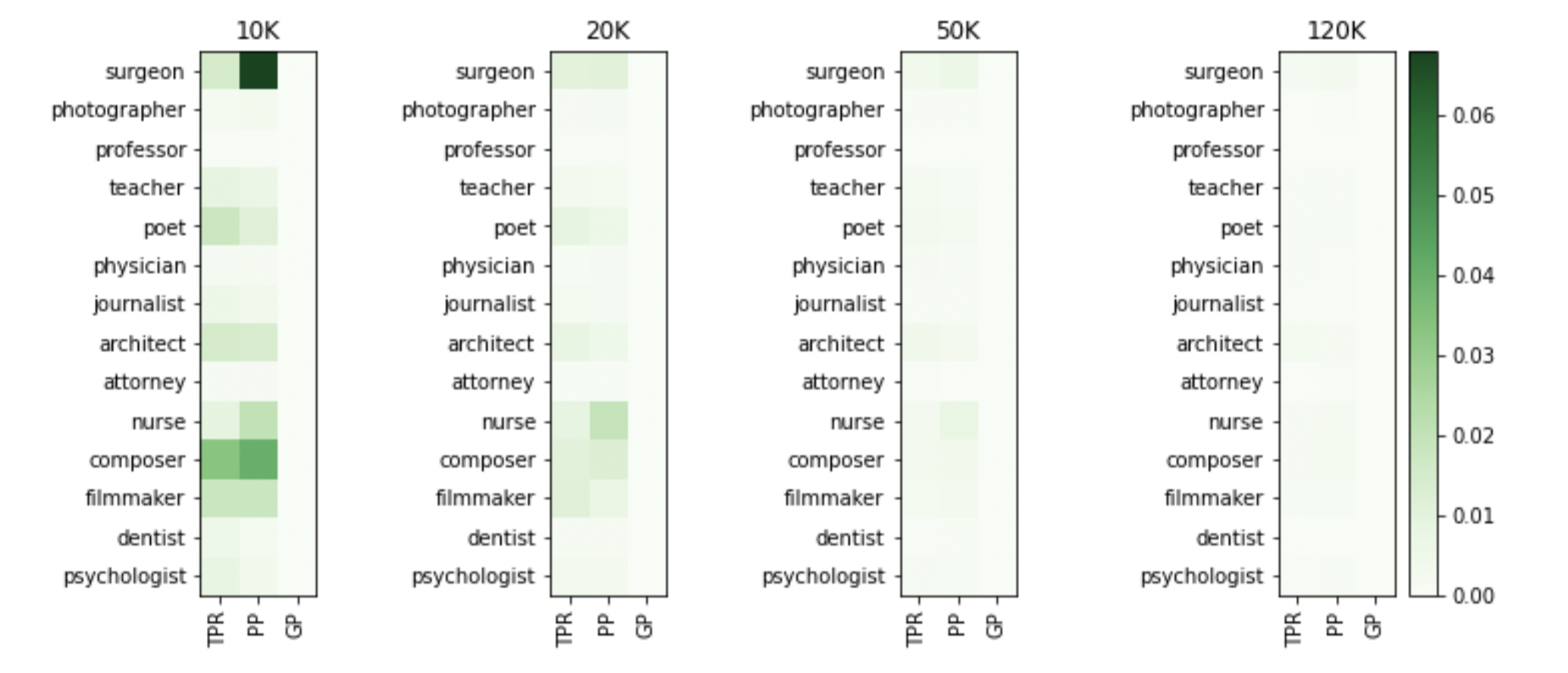}
    \caption{\textbf{Variance of TPR/PP/GP gender gap for all occupations} for model training on the classic dataset for all sample sizes. The higher the variance, the darker the green. We have 50 sampling for each sample size. We kept only professions that have at least one prediction per gender for all samplings. So we had to remove \textit{paralegal}, \textit{dj}, \textit{rapper}, \textit{pastor}, \textit{chiropractor}, \textit{software engineer}, \textit{attorney}, \textit{yoga teacher}, \textit{painter}, \textit{model}, \textit{personal trainer}, \textit{comedian}, \textit{accountant}, \textit{interior designer}, and \textit{dietitian}}
    \label{fig:TPR_PP_GP_variance}
\end{figure*}

\begin{figure*}[!h] 
    \centering
    \includegraphics[width=0.6\linewidth]{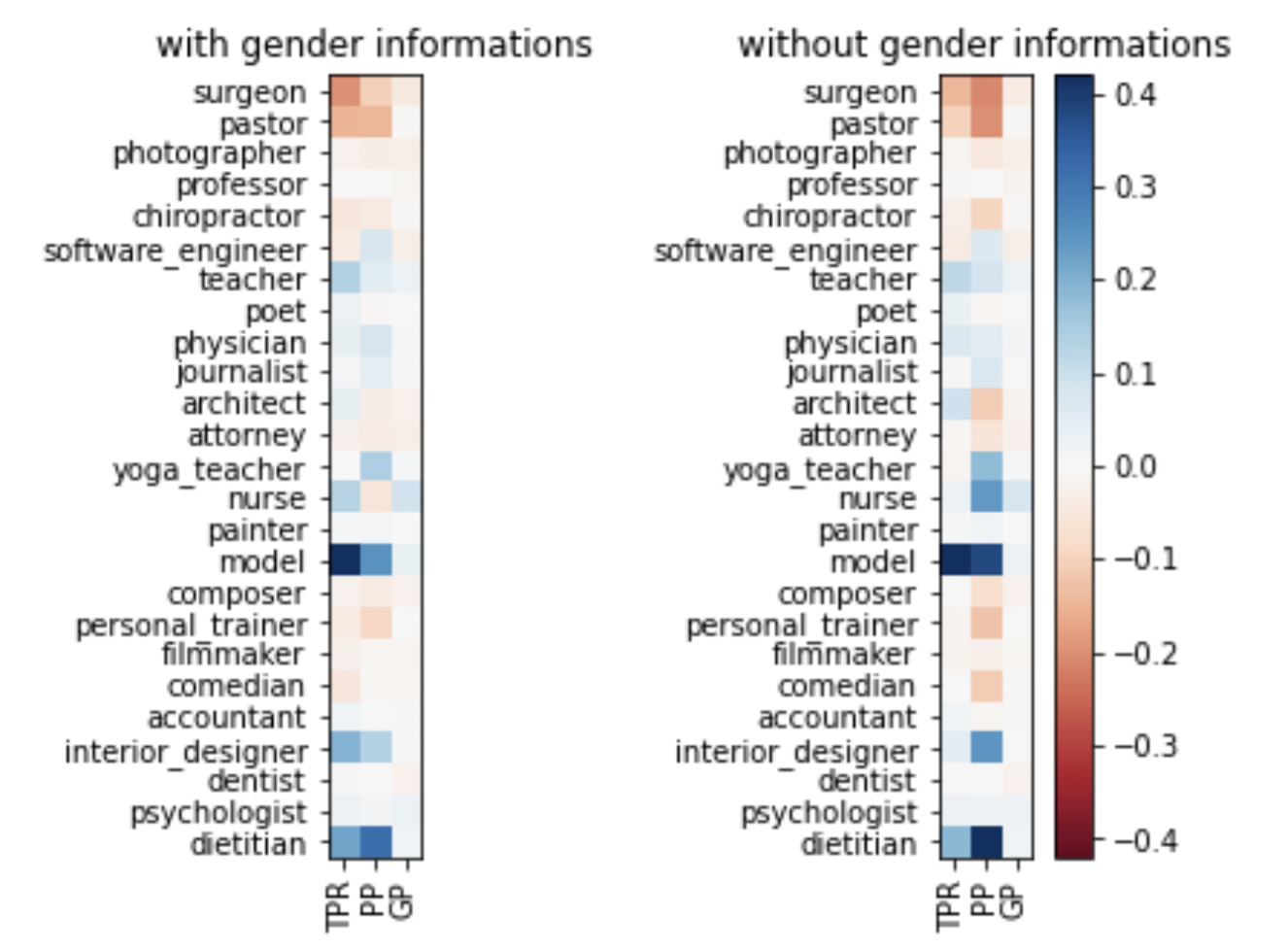}
    \caption{\textbf{Mean of TPR/PP/GP gender gap for all occupations} for model trained on 120K samplings. On the right, the model was trained on the classic dataset, and on the left, the model was trained on the dataset without gender indicators. The more it is red, the more it is biased in favor of males, the more it is blue, the more it is biased in favor of females.
    We kept only professions with more than 10 predictions per gender. So we had to remove \textit{paralegal}, \textit{dj} and \textit{rapper}.}
\label{fig:TPR_PP_GP_all_occupations}
\end{figure*}

The general results on all occupations confirm the analysis we made on the two occupations previously: \begin{enumerate}
    \item In Figure \ref{fig:TPR_PP_GP_variance}, we have more and more important deviations on the variance of the metrics as the size of the data set decreases. And that on most trades. As explained before, the GP gender gap is more stable, because it has more data. 
    \item In the first table of Figure \ref{fig:TPR_PP_GP_all_occupations} (for the classic model), the metrics give inconsistent results for several occupations: depending on the metric bias in favor of men or women for the same profession and the same model. This is particularly visible for the occupations: \textit{software engineer}, \textit{poet}, \textit{architect}, \textit{attorney}, and \textit{nurse}.
    \item By comparing the two tables in Figure \ref{fig:TPR_PP_GP_all_occupations}, we confirm that depending on the metric we are looking at, the basic debiasing technique used will not have the same effects on the bias. In several professions, we see that the bias on the TPR gender gap in favor/against women increases when on the bias on the PP gender gap decreases and vice versa. This is evident in the professions: \textit{surgeon}, \textit{pastor}, \textit{photographer}, \textit{chiropractor}, \textit{teacher}, \textit{journalist}, \textit{architect}, \textit{attorney}, \textit{nurse}, \textit{composer}, \textit{personal trainer}, \textit{comedian}, \textit{interior designer}, and \textit{dietetitan}.
\end{enumerate}

These results give us guarantees on the generalization of our analysis carried out on the two classes previously. We find the same problems with the metrics and the size of the sample, regardless of the occupation being looked at.

\section{Conclusion}


Our paper used the \textit{Bios} dataset to study the influence of the training set size on discriminatory biases. Our results shed light on new phenomena: (1) fairness metrics did not converge to stable results for small sample sizes, which precluded any conclusions about the nature of the biases;  (2) even on large training samples, the biases discovered were not always those expected and varied according to the metrics for several occupations; (3) a simple debiasing method, which consists in removing explicit gender indicators, had an unstable impact in our results depending on the metrics, though our analysis of the metrics can explain the instability. 
These results give two clear messages to data scientists who  must design NLP applications with a potential social impact. They should first be particularly careful, as the decision rules they train may have unexpected discriminatory biases.  In addition, a bias metric not only returns a score but has a strong practical meaning and may be unreliable, in particular when working with small training sets.  So multiple metrics should be considered and statistical methods to obtain the variance of the observed metrics are necessary to guarantee the fairness of a model.


\section{Limitations}

A limitation of our conclusions is that although it is necessary to use several fairness metrics to be able to properly quantify the bias, this is not enough. These metrics must be well chosen according to the context and the task being looked at. The expertise of a person working in the field is therefore always necessary to have the most complete possible interpretation of the bias. 
More specifically, the different fairness metrics measure distinct properties, and the fact that they are often incompatible has been a core part of the fair ML conversation from the beginning \cite{barocas2017fairness}. 
Thus, suggesting to choose a different metric depending on the sample size may sometimes be inappropriate, since this choice may depend on the meaning of the metric in a given application.
We must therefore be very careful and see the notion of robustness as additional necessary information and not as a replacement for the metric's meaning.

We also did not reduce the bias using advanced strategies because this paper focuses more on the analysis intended for a population closer to the law than to machine learning. In this vein, it is interesting to note that more and more tools are available to reduce bias. In particular, \cite{sikdar2022getfair}  makes it possible to reduce the bias according to several fairness metrics, therefore remaining in our logic of taking several metrics.

The main problem raised by our article comes from the fact that fairness indices are not stable when they are calculated. We should consider them as random variables and look at their law. The first step is to look at the mean and the variance as done in this paper but having the full distribution would be more interesting. Works that compute the asymptotic law can be taken as an example like \cite{ji2020can, besse2022survey}.

\section*{Ethics Statement}

Natural Language Processing is gaining a considerable amount of attention these days and it is extremely important to evaluate how NLP datasets will impact the gender bias when used to train models that will be used in the real world. This work uses different experiments and fairness metrics to shed light on the shortcomings of these metrics with respect to gender bias made by ML algorithms on textual data.  We believe that transmitting knowledge from research to industry on a subject like fairness is essential to make the field of ML more ethical. Hence, this work focuses on issues that most affect the industrial ML landscape and contains a clear message to them on how they should change their current practices.


\section*{Acknowledgements}
This research was funded by the AI (Artificial Intelligence) Interdisciplinary Institute ANITI (Artificial and Natural InTelligence Institute.), which is funded by the French ‘Investing for the Future– PIA3’ program under the Grant agreement ANR-19-PI3A-0004.

\bibliography{anthology,custom}
\bibliographystyle{acl_natbib}

\appendix

\section{Mathematical intuition}
\label{apdx:mathintuition}
\paragraph{Intuition} Let $\hat{Y}$ and $Y$ be the predicted and the true target labels, respectively. Let $G$ be a random variable representing the binary gender and let $n$ be the number of all individuals.
We can write the estimators of Group Parity, True Positive Rate, and Predictive Parity metrics like this:
\begin{align*}
    \hat{GP}_{g,y} &= \frac{\sum_{i=1}^n1_{\{ \hat{Y}_i = y \ \cap \ G_i = g\}}}{\sum_{i=1}^n1_{\{G_i = g\}}} \\
    \hat{TPR}_{g,y} &= \frac{\sum_{i=1}^n1_{\{ \hat{Y}_i = y \ \cap \ Y_i=y  \ \cap \ G_i = g\}}}{\sum_{i=1}^n1_{\{Y_i=y \ \cap \ G_i = g\}}} \\
    \hat{PP}_{g,y} &= \frac{\sum_{i=1}^n1_{\{ \hat{Y}_i = y \ \cap \ Y_i=y  \ \cap \ G_i = g\}}}{\sum_{i=1}^n1_{\{\hat{Y}_i = y \ \cap \ G_i = g\}}}
\end{align*}

We set $A = \{ \hat{Y}_i = y \ \cap \ G_i = g\}$ and $B = \{ Y_i = y \}$. 
By definition, $\#(A \ \cap \ B) \leq \#A$ where $\#$ is the cardinal of the set. So we have $\#\{ \hat{Y}_i = y \ \cap \ Y_i = y \ \cap \ G_i = g\} \leq \# \{ \hat{Y}_i = y \ \cap \ G_i = g\}, \forall i=1,...,n$.

We can define $n_{GP}$, $n_{TPR}$, $n_{PP}$  the number of individuals respectively looked by the estimator of Group Parity, True Positive Rate, and Predictive Parity metrics and we have: 
\begin{align*}
    n_{GP} &= \sum_{i=1}^n\# (\{ \hat{Y}_i = y \ \cap \ G_i = g\} \ \cap \ \{ G_i = g\} ) \\
    &= \sum_{i=1}^n\#\{ \hat{Y}_i = y \ \cap \ G_i = g\}, \\
n_{TPR} &= \sum_{i=1}^n\# (\{ \hat{Y}_i = y \ \cap \ Y_i=y \ \cap \ G_i = g\} \\
& \cap \ \{ Y_i=y \ \cap \ G_i = g\} ) \\
&= \sum_{i=1}^n\#\{ \hat{Y}_i = y \ \cap \ Y_i=y \ \cap \ G_i = g\}, \\
n_{TPR} &= \sum_{i=1}^n\# (\{ \hat{Y}_i = y \ \cap \ Y_i=y \ \cap \ G_i = g\} \\
& \cap \ \{ \hat{Y}_i = y \ \cap \ G_i = g\} ) \\
&= \sum_{i=1}^n\#\{ \hat{Y}_i = y \ \cap \ Y_i=y \ \cap \ G_i = g\}.
\end{align*}
Then : $n_{TPR} = n_{PP} \leq n_{GP}$. \\

\section{Additional figure}
\begin{figure*}[!h] 
    \centering
\includegraphics[width=1\linewidth]{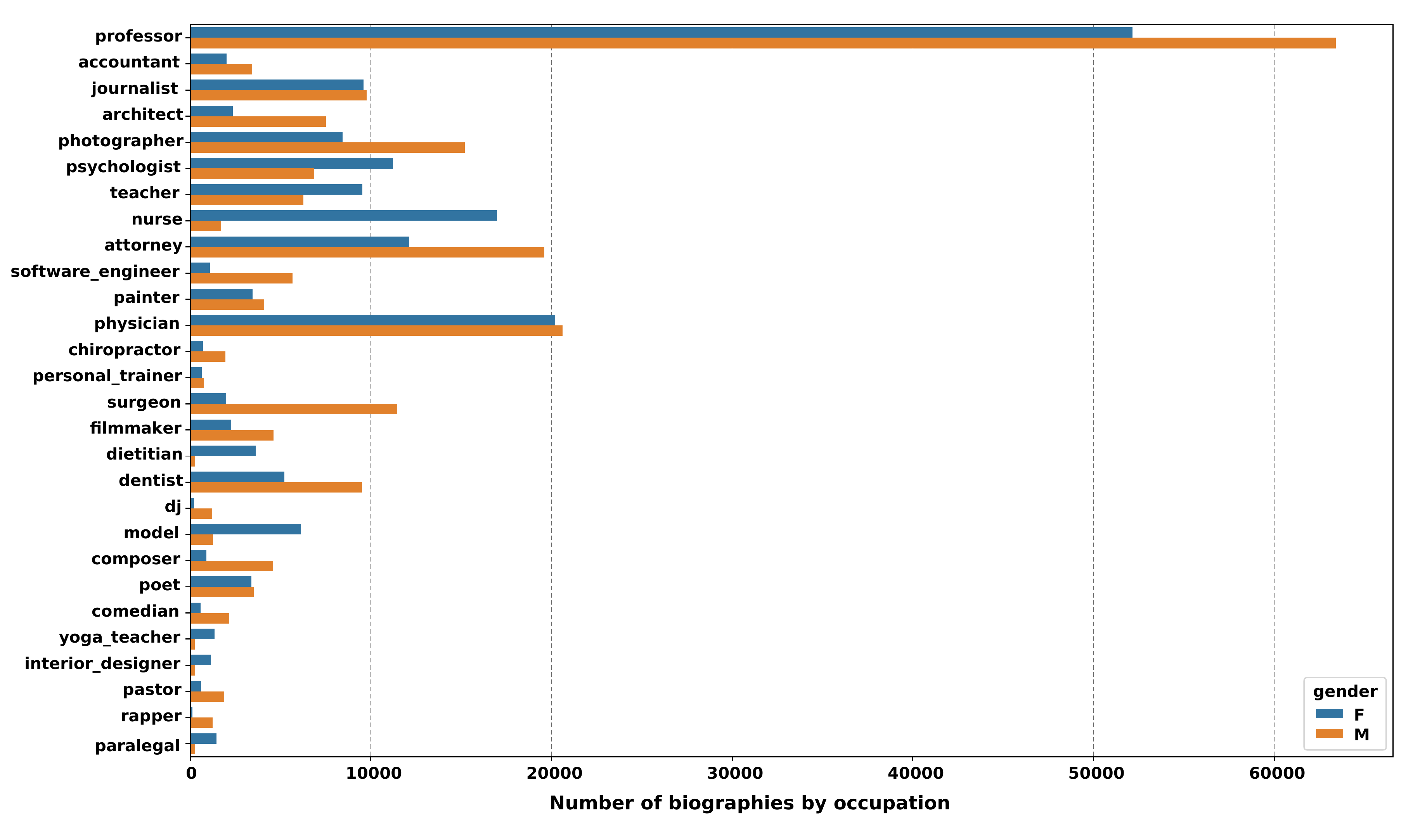}
    \caption{Number of biographies for each occupation by gender on the total dataset}
    \label{fig:distribution}
\end{figure*}
\end{document}